# Towards an Interoperable Ecosystem of AI and LT Platforms:
# A Roadmap for the Implementation of Different Levels of Interoperability

**Georg Rehm[1], Dimitrios Galanis[2], Penny Labropoulou[2], Stelios Piperidis[2], Martin Welß[3], Ricardo Usbeck[3], Joachim Köhler[3], Miltos Deligiannis[2], Katerina Gkirtzou[2], Johannes Fischer[4], Christian Chiarcos[5], Nils Feldhus[1], Julián Moreno-Schneider[1], Florian Kintzel[1], Elena Montiel[6], Víctor Rodríguez Doncel[6], John P. McCrae[7], David Laqua[3], Irina Patricia Theile[3], Christian Dittmar[4], Kalina Bontcheva[8], Ian Roberts[8], Andrejs Vasiļjevs[9], Andis Lagzdiņš[9]**

[1] DFKI GmbH, Germany • [2] ILSP/Athena RC, Greece • [3] Fraunhofer IAIS, Germany • [4] Fraunhofer IIS, Germany •
[5] Goethe University Frankfurt, Germany • [6] Universidad Politécnica de Madrid, Spain •
[7] National University of Ireland Galway, Ireland • [8] University of Sheffield, UK • [9] Tilde, Latvia

Corresponding author: Georg Rehm – georg.rehm@dfki.de

## Abstract

With regard to the wider area of AI/LT platform interoperability, we concentrate on two core aspects: (1) cross-platform search and discovery of resources and services; (2) composition of cross-platform service workflows. We devise five different levels (of increasing complexity) of platform interoperability that we suggest to implement in a wider federation of AI/LT platforms. We illustrate the approach using the five emerging AI/LT platforms AI4EU, ELG, Lynx, QURATOR and SPEAKER.

## 1. Introduction

Due to recent breakthroughs in deep neural networks, artificial intelligence has been increasingly ubiquitous in the society and media. AI is now widely considered a continuous game-changer in every technology sector. While critical aspects need to be carefully considered, AI is perceived to be a big opportunity for many societal and economical challenges. As a prerequisite, a large number of AI platforms are currently under development, both on the national level, supported through local funding programmes, and on the international level, supported by the European Union. In addition to publicly-supported endeavours, many companies have been developing their own clouds to offer their respective services or products in their targeted sectors (including legal, finance, health etc.). Positioned orthogonally to these verticals, Language Technology (LT) platforms typically offer domain-independent, sometimes domain-specific, services for the analysis or production of written or spoken language. LT platforms can be conceptualised as language-centric AI platforms: they use AI methods to implement their functionalities. Various European LT platforms exist, both commercial and non-commercial, including large-scale research infrastructures.

The enormous fragmentation of the European AI and LT landscape is a challenge and bottleneck when it comes to the identification of synergies, market capitalisation as well as boosting technology adoption and uptake (Rehm et al., 2020c). The fragmentation also relates to the number and heterogeneity of AI/LT platforms. If we do not make sure that all these platforms are able to exchange information, data and services, their increasing proliferation will further contribute to the fragmentation rather than solve it. This can be achieved by agreeing upon and implementing standardised ways of exchanging repository entries and other types of metadata or functional services, or enabling multi-platform and multi-vendor service workflows, benefitting from their respective unique offerings. Only by discussing and agreeing upon standards as well as technical and operational concepts for AI/LT platform interoperability, can we benefit from the highly fragmented landscape and its specialised platforms. This paper takes a few initial steps, which we demonstrate primarily using the two platforms AI4EU and ELG (European Language Grid) but also including QURATOR, Lynx and SPEAKER. These platforms are introduced in Section 2, where we also compare their architectures. Section 3 introduces requirements and prerequisites for platform interoperability, including shared semantics as well as legal and operational interoperability, followed by a description of five levels of platform interoperability that exhibit an increasing level of conceptual complexity. Section 4 summarises the paper and presents next steps. We contribute to the challenge of platform interoperability by identifying this topic as a crucial common development target and by suggesting a roadmap for the implementation of different levels of interoperability.

## 2. The Platforms

In the following, we describe the platforms AI4EU (Section 2.1), ELG (Section 2.2), QURATOR (Section 2.3), Lynx (Section 2.4) and SPEAKER (Section 2.5).

### 2.1. AI4EU

In January 2019, the AI4EU consortium with more than 80 partners started its work to build the first European AI on-demand platform. The main goals are: the creation and support of a large European ecosystem to facilitate collaboration between all European AI actors (scientists, entrepreneurs, SMEs, industries, funding agencies, citizens etc.); the design of a European AI on-demand platform to share AI resources produced in European projects, including high-level services, expertise in research and innovation, components and data sets, high-powered computing resources and access to seed funding for innovative projects; the implementation of industry-led pilots, which demonstrate the platform's capabilities to enable real applications and foster innovation; research in five key areas (explainable AI, physical AI, verifiable AI, collaborative AI, integrative AI); the creation of a European ethical observatory to ensure that European AI projects adhere to high ethical, legal, and socio-economical standards; the preparation of a Strategic Research Innovation Agenda for Europe.

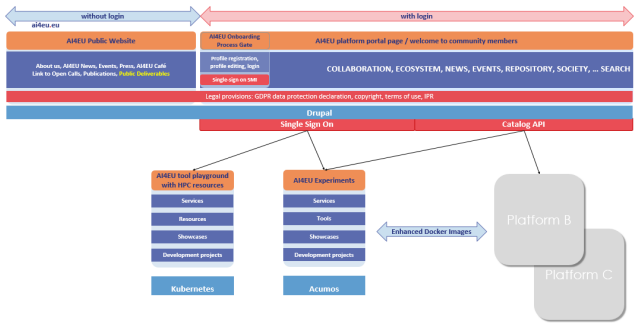

Figure 1: AI4EU logical structure

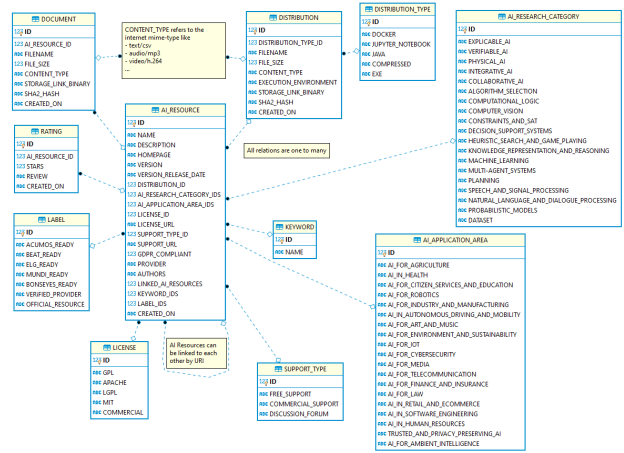

Figure 2: AI4EU metadata model

The AI4EU platform consists of several subsystems. In this paper, we focus on the AI4EU Repository and AI4EU Experiments, which are at the core of all interoperability topics. The repository exposes the Catalog API, which is based on the AI4EU metadata model, in the center of which is the AI resource: this can be any relevant entity like trained models, data sets, tools for symbolic AI, tools to build AI pipelines etc. AI resources can be linked to each other, e. g., a trained model could be linked to the data set used for training. The license information is mandatory demonstrating the emphasis on lawful reuse of resources. Documents, pictures and binary artefacts can be associated with a resource. However, AI resources cannot be combined or worked with in the repository itself. That leads us to the AI4EU Experiments subsystem, which enables the quick and visual composition of AI solutions using tools with published, well-known interfaces. These solutions can be training or production pipelines or pipelines to check or verify models. The subsystem enables easily to connect tools to data sets via databrokers or datastreams. It includes tools and models for symbolic AI, ethical AI and verifiable AI, and allows for collaboration and feedback (discussion, ratings, workgroups). It also supports mixed teams, e. g., with business users and external AI experts to bootstrap AI adoption in SMEs. To combine tools to runnable pipelines, the expected format of an AI resource is an enhanced Docker container, which (1) contains a license file for the resource; (2) includes a self-contained protobuf[1] specification of the service, defining all input and output data structures; (3) exposes the above service using gRPC.[2] Protobuf and gRPC are both open source and programming language-neutral and, thus, a solid foundation for interoperability, especially when combined with Docker.

**Interoperability is addressed at the following levels:** (1) AI4EU supports the bidirectional exchange of metadata of AI resources, i. e., to send and receive catalog entries. Since AI4EU is prepared to connect with other platforms, it takes the approach of focussing the metadata on the least common denominator. This docking point is the Catalog API. (2) To contribute to a distributed search across several platforms, AI4EU provides a search API. It accepts remote queries, executes them on the catalog and returns a list of matches from the AI4EU repository. (3) The Docker container format used in AI4EU Experiments.

### 2.2. European Language Grid (ELG)

Multilingualism and cross-lingual communication in Europe can only be enabled through Language Technologies (LTs) (Rehm et al., 2016). The European LT landscape is fragmented (Vasiljevs et al., 2019), holding back its impact. Another crucial issue is that many languages are under-resourced and, thus, in danger of digital extinction (Rehm and Uszkoreit, 2012; Kornai, 2013; Rehm et al., 2014). There is an enormous need for an European LT platform as a unifying umbrella (Rehm and Uszkoreit, 2013; Rehm et al., 2016; STOA, 2017; Rehm, 2017; Rehm and Hegele, 2018; European Parliament, 2018; Rehm et al., 2020c).

The project European Language Grid (2019-2021) attempts to establish the primary platform and marketplace for the European LT community, both industry and research (Rehm et al., 2020a). This scalable cloud platform will provide access to hundreds of LTs for all European languages, including running services as well as data sets. ELG will enable the European LT community to upload their technologies and data sets, to deploy them, and to connect with other resources. ELG caters for *commercial* and *non-commercial LTs* (i. e., LTs with a high Technology Readiness Level, TRL), both *functional* (processing and generation, written and spoken) and *non-functional* (data sets etc.). The platform has a user interface, backend components and APIs. Functional services are made available through containerisation and by wrapping them with the ELG LT Service API.[3] These services, provided initially by members of the ELG consortium and ultimately by many external partners, can be used through APIs or the web UI (Figure 3). The *base infrastructure* is operated on a Kubernetes[4] cluster in the data centre of a Berlin-based cloud provider. All infrastructural components of the three layers run in this cluster as Docker containers. They are built with robust, scalable, reliable and widely used technologies and frameworks, e. g., Django, Drupal, ReactJS, AngularJS.

---

[1]https://developers.google.com/protocol-buffers
[2]https://grpc.io
[3]https://gitlab.com/european-language-grid/platform/
[4]https://kubernetes.io

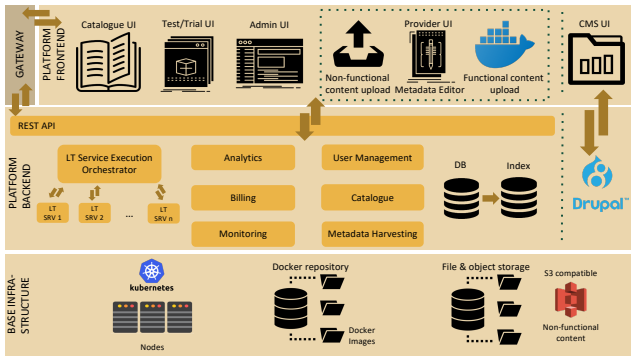

Figure 3: Technical architecture of the ELG

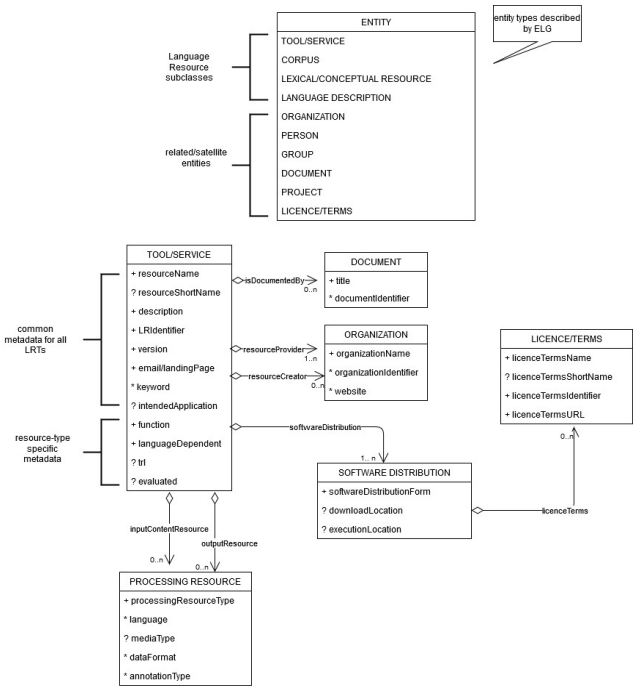

Figure 4: Excerpt of the ELG metadata schema

The *backend* contains the catalogue, i. e., the list of metadata records of services, resources, organisations (e. g., companies, universities, research centres), service types, languages etc. Stakeholders will be able to register themselves, ensuring increased reach and visibility. Users can filter and search for organisations, services, data sets and more, by language, service type, domain, and country. Functionalities are offered via REST services. Metadata records are stored in PostgreSQL and ElasticSearch. The LT Service Execution Server offers a common REST API. The *frontend* consists of UIs for different user types, e. g., LT providers, buyers and system administrators. These include catalogue UIs, test UIs for functional services, provider UIs for uploading/registering services etc.

ELG uses Docker containers to encapsulate all components, settings and libraries of an individual LT service in one self-contained unit. Docker images can be built locally by their developers and ingested into the ELG, where they can be started, terminated and scaled out on demand. Containers can be also replaced easily.

Kubernetes is used for container orchestration. It decides autonomously how many replicas of an LT service are needed at any given point in time.[5] The integration of a service into the ELG currently consists of six steps: (1) adapt the service to the ELG API; (2) create a Docker image; (3) push the Docker image into a registry (e. g., ELG Gitlab); (4) request, from the ELG administrators, a Kubernetes namespace, in case of a proprietary service with restricted access; (5) deploy the service by creating a Kubernetes config file; (6) add the service to the ELG catalogue by providing the metadata. For some of the approx. 175 services currently in the ELG, this process took a few days, for others, only a few hours. Our goal is to bring this effort down to a minimum, at least for the most common cases.

The ELG metadata schema (Labropoulou et al., 2020) supports discovery and operation for humans and machines. It describes Language Resources and Technologies (LRTs) and related entities (organizations, persons, projects, etc.; Figure 4). The schema is organised around three concepts:

---

[5]For autoscaling and scale-to-zero functionalities, ELG uses Knative (https://cloud.google.com/knative).

*resource type* (tool or service, corpus, lexical or conceptual resource, language description), *media type* (text, audio, video, image) and *distribution*, i. e., the physical form of the resource (e. g., software distributed as web services, source or binary code). Administrative and descriptive metadata (e. g., identification, contact, licensing information, etc.) are common to all LRTs, while technical metadata differ across resource/media type and distributions.

**Interoperability is addressed at the following levels:** (1) exchange of metadata records from and to other, external catalogues: the schema exploits an RDF/OWL ontology (McCrae et al., 2015) with links to widespread vocabularies and ontologies and the possibility to be further enriched with those of collaborating initiatives; (2) interoperability across resource types, supporting the automatic match of (a) candidate resources that can be combined together to form a workflow (e. g., matching input and output formats of tools to create pipelines, models of a specific type with tools that can utilize them), and (b) data resources with functional services that can be used for their processing (e. g., an English NER tool with English data sets etc.).

### 2.3. QURATOR: Curation Technologies

Online content has recently gained immense importance in many areas of society. Some of the challenges include better support and smarter technologies for content curators who are exposed to an ever increasing stream of heterogeneous information they need to process, e. g., knowledge workers in libraries digitize archives, add metadata and publish them online, journalists need to continuously stay up to date on their current topic of investigation. Many work environments would benefit immensely from technologies that support content curators (Rehm and Sasaki, 2015).

The QURATOR consortium consists of ten partners from industry and research (Rehm et al., 2020b). The project de-velops a curation technology platform, which is also being populated with services, simplifying and accelerating the curation of content (Bourgonje et al., 2016a; Rehm et al., 2019a; Schneider and Rehm, 2018a; Schneider and Rehm, 2018b). The project develops, evaluates and integrates services for preprocessing, analyzing and generating content, spanning use cases from the sectors of culture, media, health and industry. To process and transform incoming data, text or multimedia streams into device-adapted, publishable content, various groups of components, services and technologies are applied. These include adapters to data, content and knowledge sources, as well as infrastructural tools and AI methods for the acquisition, analysis and generation of content. All these different technologies are combined into pilots and prototypes for selected use case.

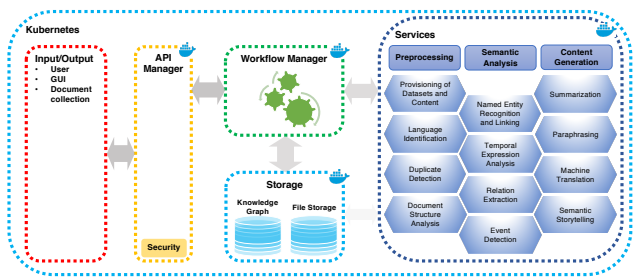

Figure 5: Technical architecture of the QURATOR platform

The QURATOR platform (Figure 5) is designed together with all partners who also contribute services, which can be divided into three broad groups: (1) *Preprocessing* encompasses services for obtaining and processing information from different content sources so that they can be used in the platform and integrated into other services (Schneider et al., 2018), e. g., provisioning content, language and duplicate detection as well as document structure recognition. (2) *Semantic analysis services* process a document and add information in the form of annotations, e. g., NER, temporal expression analysis, relation extraction, event detection, fake news as well as discourse analysis (Bourgonje et al., 2016b; Srivastava et al., 2016; Rehm et al., 2017b; Ostendorff et al., 2019). (3) *Content generation services* enable the creation of a new piece of content, e. g., summarization, paraphrasing, and semantic storytelling (Rehm et al., 2019c; Rehm et al., 2018; Moreno-Schneider et al., 2017; Rehm et al., 2017a; Schneider et al., 2017; Schneider et al., 2016).

**Interoperability is addressed at the following levels:** Since the QURATOR platform is a closed ecosystem, the platform can be thought of as an experimental toolbox with services customised by the partners for their own use cases. As the platform is used only by the QURATOR partners, it does not contain a catalogue or any kind or structured metadata. However, two of the ten QURATOR projects have a focus on service composition and workflows with prototypical implementations under development (Moreno-Schneider et al., 2020a), using NIF as a joint annotation format (Hellmann et al., 2013).

### 2.4. Lynx: Legal Knowledge Graph Platform

The project Lynx produces a multilingual Legal Knowledge Graph (LKG), in which data sources from different jurisdictions, languages and orders are aggregated and interlinked by a collection of analysis and curation services. Lynx aims to facilitate compliance of SMEs and other companies in internationalisation processes, leveraging European legal and regulatory open data duly interlinked and offered through cross-sectorial, cross-lingual services. The platform is tested in three pilots that develop solutions for legal compliance, regulatory regimes and compliance, where legal provisions, case law, administrative resolutions, and expert literature are interlinked, analysed, and compared to inform strategies for legal practice.

The platform (Figure 6) focuses upon three main components: (1) semantic services for the extraction of information from large and heterogeneous sets of documents; (2) the LKG (Montiel-Ponsoda and Rodríguez-Doncel, 2018; Schneider and Rehm, 2018a; Martín-Chozas et al., 2019) stores linguistic and legal information from documents; (3) the workflow manager realises complex use cases.

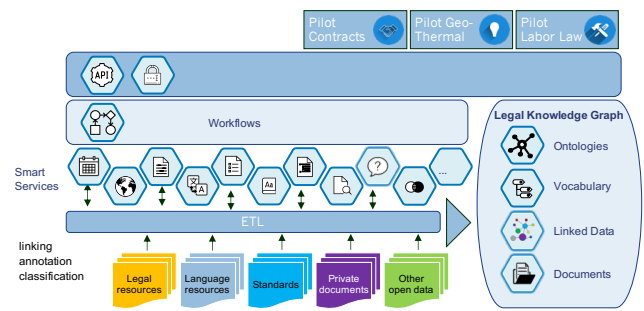

Figure 6: The Lynx technology platform

The platform's microservice architecture is a variant of the service-oriented architecture (SOA), in which an application is structured as a collection of loosely coupled services. It uses Docker containers hosted and managed through OpenShift, a containerisation software built on top of Kubernetes.[6] Services communicate through REST APIs. The platform includes a heterogeneous set of services (Rehm et al., 2019b).[7] Some of the services make use of others, some extract or annotate information, while others operate on full documents, yet others provide a user interface. The Document Manager provides the storage and annotation of documents with an emphasis on keeping them synchronized, providing read and write access, as well as updates of documents and annotations. It can be queried in terms of annotations and documents, through REST APIs. The interface includes a set of create, read, update, and delete APIs to manage collections, documents and annotations. The orchestration and execution of services involved in more complex tasks is addressed by a Workflow Manager. It defines combinations of services as workflows (Moreno-Schneider et al., 2020b; Bourgonje et al., 2016a; Schneider and Rehm, 2018a; Schneider and Rehm, 2018b). Workflows are described using BPMN and executed using Camunda.[8]

**Interoperability is addressed at the following levels:** Like all previously described platforms, the Lynx platform is based on microservices orchestrated as containers. Like the QURATOR platform, the Lynx platform does not contain a structured catalogue with metadata entries other than Open API descriptions, because some services have restricted access and, so far, are only used by the project partners. While the QURATOR platform is populated with a large variety of services, the development of the domain-specific Lynx services is primarily driven by three focused use cases. The Lynx platform includes a workflow manager. Lynx defines an RDF-based data model, which reuses NIF (Hellmann et al., 2013), ELI (European Legislation Identifier) metadata elements and other standard specifications. A SHACL-based validator grants conformance and favours interoperability.

---
[6]https://www.openshift.com
[7]http://lynx-project.eu/doc/api/
[8]https://camunda.com

### 2.5. SPEAKER

The SPEAKER project develops a B2B conversational agent platform "Made in Germany". A secondary aim is the creation of a vivid ecosystem. Numerous partners, such as large industrial companies, SMEs, start-ups and research partners ensure the project's practical relevance, as well as academic excellence. Industry expressed a strong demand for a speech assistant platform that can accommodate specific application scenarios. These use cases comprise, e. g., an automated speech recognition (ASR) component that can be adapted to recognize technical terms or the unification of company-internal knowledge graphs using NLP.

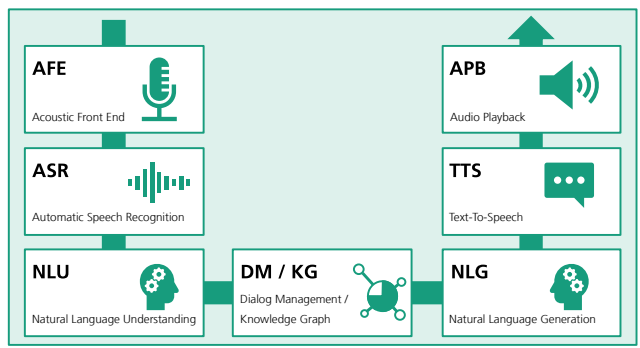

Figure 7: Flexible workflow components in SPEAKER

The speech solutions developed by the large technology providers based on other continents do not offer the required customizability nor do they comply with GDPR. Thus, they do not meet the data protection standards required by many of the SPEAKER industry partners. In many use cases, data that needs to be handled by a conversational agent is either sensitive (e. g., medical records) or company secrets, the confidentiality of which must not be jeopardized.

The platform will comprise core modules such as AFE, ASR, NLU, DM/KG, NLG, TTS and APB. These can be combined to implement complete B2B voice assistant applications (Figure 7). Each module can also be deployed individually, customized to the targeted use case. Platform interoperability will be investigated during the lifetime of the project. The SPEAKER partners have the necessary know-how and expertise (Usbeck, 2014; Both et al., 2014; Singh et al., 2018; Shet et al., 2019; Govalkar et al., 2019; Fischer et al., 2016; Chakrabarty and Habets, 2019), enabling them to develop this flexible and scalable platform.

**Interoperability is addressed at the following levels:** SPEAKER will provide a modular, customizable platform based on mature, existing components. It is intended to implement the industry partners' use cases in a close to production ready fashion. Thus, high quality and reliable services with the additional privacy features are required. SPEAKER will investigate interfaces to other platforms in order to facilitate interoperability. SPEAKER is less open to ensure a high level of trust and data privacy. In contrast to QURATOR and Lynx, it will have a structured service catalog for self-servicing. SPEAKER will offer an orchestration component to enable the flexible composition of voice assistants. Services will be containerised using Docker and hence be pluggable into on-premise computing landscapes.

### 2.6. Common Aspects and Functionalities

The five platforms share several common aspects but also differ substantially with regard to other dimensions and requirements. Table 1 provides a comparison. While AI4EU caters for AI at large, ELG concentrates on LT, i. e., language-centric AI. Lynx, QURATOR and SPEAKER focus upon specific domains and application areas within LT. AI4EU and ELG are community-driven, open platforms through which third parties can make available services or resources, while the other three are closed, i. e., populated by their respective project consortia with the goal of commercial exploitation. All platforms make use of microservices and orchestrate their containers through base infrastructures that provide mechanisms for scaling. Structured repositories of services and resources are maintained in AI4EU, ELG and SPEAKER; all platforms with a repository also have a graphical user interface enabling search and discovery of resources. Workflows are at least partially addressed in all platforms except ELG; however, it is planned to evaluate if the QURATOR approach can be integrated into the ELG platform (Moreno-Schneider et al., 2020a). Table 1 also includes ranges with regard to the targeted Technology Readiness Level (TRL) of the platforms and their services.[9] The individual TRLs indicate the range between a rather experimental and a more production-ready stage of the platform initiatives and their services.

Technically and conceptually, interoperability between these or other AI/LT platforms can be addressed with regard to the repository layer, the API layer, the functional service layer (workflows) or the computation layer.

### 3. Platform Interoperability

Platform interoperability can be achieved with regard to various different aspects. We concentrate on two that are inspired by the heterogeneous European landscape: (1) cross-platform search and discovery of resources and (2) composition of cross-platform workflows. The broad and robust implementation of these two feature sets makes it possible to use the search functionality of platform A with specific criteria and to receive matches, if any, from all platforms attached to platform A. The cross-platform composition of service workflows enables putting together distributed processing pipelines that make use of REST services hosted on different platforms. We can even think of more complex service development scenarios in which we, e. g., take a data set, hosted on ELG, ingest it into the AI4EU Experiments instance, train a new model and move the resulting Kubernetes artefact back into ELG, describing it with metadata, making it available to all platforms.

---
[9] https://en.wikipedia.org/wiki/Technology_readiness_level

| | Scope | Domain-specific | Open vs. Closed | Infrastructure | Structured Catalogue | Functional Microservices | Workflows possible | Targeted TRL of ... platform | services |
|---|---|---|---|---|---|---|---|---|---|
| **AI4EU** → https://www.ai4eu.eu – Runtime: 01/2019–12/2021 | Europe | no (AI at large) | Open | Kubernetes, Acumos, Drupal | yes | yes | yes | 7-9 | 6-9 |
| **ELG** → https://www.european-language-grid.eu – Runtime: 01/2019–12/2021 | Europe | no (LT at large) | Open | Kubernetes, Drupal | yes | yes | no | 7-9 | 5-9 |
| **Lynx** → http://lynx-project.eu – Runtime: 12/2017–11/2020 | Europe | Legal domain | Closed | OpenShift | no | yes | yes | 7-8 | 6-8 |
| **QURATOR** → https://qurator.ai – Runtime: 11/2018–10/2021 | Germany | Curation services | Closed | Kubernetes | no | yes | partially | 4-6 | 3-8 |
| **SPEAKER** → https://www.speaker.fraunhofer.de – Runtime: 04/2020–03/2023 | Germany | Voice Assistants | Closed | Kubernetes | yes | yes | yes | 8-9 | 8-9 |

Table 1: Central characteristics of selected emerging European AI/LT platforms

Before we provide more details on the five levels of platform interoperability (Sections 3.2 to 3.6), we discuss the benefits of using a shared semantic space for achieving interoperability; we also describe a solution for creating it in the form of a reference model acting as a bridge between the metadata schemas of the different platforms and that may also provide interoperability on the level of exchange formats or annotations (Section 3.1). Finally, Section 3.7 discusses the aspect of legal and operational interoperability.

### 3.1. Shared Semantic Space

For the more advanced levels of platform interoperability (Level 2 and upwards), a shared semantic space is needed as a joint, ontologically grounded and machine-readable vocabulary, into which all platform-specific concepts and terminologies can be mapped so that abstract conceptualisations originating in a platform, e. g., names of service categories or specific annotation labels, can be interpreted. Such a shared semantic space explicitly represents knowledge about various different aspects, including, among others: (1) categories of resources including different types of data resources (data set, corpus, lexicon, terminology, language model, etc.) and different types of tools and functional services (NER, parser, image classifier, facial expression detector, etc.); (2) abstract descriptions of the I/O requirements of tools and services (data formats, languages, modalities etc.); (3) attributes and values used in specific annotation formats and tagsets including metadata about annotation formats themselves.

As a first step, interoperability can be achieved by mapping two schemas onto each other and creating converters. However, such an approach does not scale because we would need to create new converters for each new platform "attached" to this federation of platforms. In contrast, the proposed shared semantic space can function as a reference model that is able to represent all crucial information typically contained in the respective platform-specific metadata scheme. Alternatively, all platforms should adhere to a joint RDF/OWL ontology for their semantic metadata. On top of the domain-independent semantic categories, there is the challenge of representing domain-specific terms and concepts. Even for general categories, communities tend to use different terms for similar concepts, which makes the adoption of a single joint ontology an almost impossible task (Labropoulou et al., 2018).

This is not the first attempt at such a shared semantic space. Previous experience does, however, show, that centralized repositories for data categories may face long-term sustainability issues (Langendoen, 2019; Warburton and Wright, 2019). As an alternative, one may consider to follow a Linked Data approach, where concepts and definitions of different providers are defined in a self-contained formal model, e. g., an ontology, and subsequently refer to vocabularies or reference concepts developed in a distributed fashion by the broader community.

This approach can be exemplified by the Ontologies of Linguistic Annotation (Chiarcos, 2008; Chiarcos and Sukhareva, 2015), a central hub for linguistic annotation terminology in the web of data. OLiA was designed for mediating between various terminology repositories on the one hand and annotated resources (i. e., their annotation schemes), on the other. Four different types of ontologies are distinguished (Fig. 8): (1) The OLiA Reference Model is an OWL ontology that specifies the common terminology that different annotation schemes can refer to. (2) Multiple OLiA Annotation Models formalize annotation schemes and tagsets. Fig. 8 illustrates this with an annotation model developed as part of the Korean NLP2RDF stack (Hahm et al., 2012). (3) For every annotation model, a linking model defines subclass-relationships between concepts in the annotation model and the reference model. Linking models are interpretations of annotation model concepts and properties in terms of the reference model. (4) Similarly, other community-maintained vocabularies are linked with OLiA, e. g., the CLARIN Concept Registry (Chiarcos et al., 2020). OLiA was developed as part of an infrastructure for the sustainable maintenance of linguistic resources (Wörner et al., 2006; Schmidt et al., 2006; Rehm et al., 2008b; Witt et al., 2009; Rehm et al., 2009). Its field of application included the formalization of annotation schemes and concept-based querying over heterogeneously annotated corpora (Rehm et al., 2008a). As several institutions and resources from various disciplines were involved, no holistic annotation standard could be enforced onto the contributors.

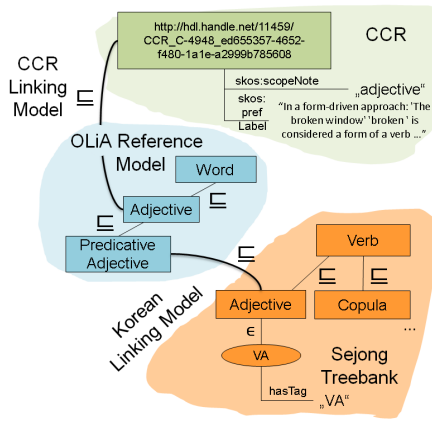

Figure 8: Modular OLiA ontologies

### 3.2. Level 1: Simple Cross-Platform Search through Public APIs

The first level of platform interoperability relates to simple cross-platform search through publicly available search and discovery APIs for resources offered by the platforms, i. e., data sets, functional services, tools, workflows, lists of organisations etc. Making the search API of a platform available to third parties enables other platform providers to integrate it in their own search facilities and, thus, to include the resources of this platform into their search space. This way, a query would return matches from all platforms. Search results need to show only minimal metadata and redirect the user to the original platform. Realising this level of interoperability requires only a limited amount of discussion and agreement between the platform operators with regard to metadata schemes, their semantics or the data format returned by the search API.

### 3.3. Level 2: Complex Cross-Platform Search through the Exchange of Metadata Records

One disadvantage of Level 1 interoperability relates to the fact that the user experience will be rather lacking because the search results retrieved from external platforms are difficult to integrate and aggregate into the search results of the local platform due to the lack of a shared semantic space; ranking search results is equally difficult. Level 2 foresees either aligning all platforms involved in such a federation of platforms along a shared semantic space that explicitly provides semantics for the metadata fields and their values, or agreeing upon the same metadata scheme or at least upon a certain (obligatory) subset (Labropoulou et al., 2020; McCrae et al., 2015). Such a more detailed, semantics-driven approach enables more efficient and more user-friendly search results from multiple platforms that can be visually aggregated and also easily ranked. The actual search can be performed through publicly available APIs but returned objects would be semantically richer. Alternatively, the metadata records of external repositories can be harvested using standard protocols such as OAI-PMH, which allow the construction of a master index out of decentralised inventories (Piperidis, 2012). A known issue that needs to be addressed using such an approach involves the detection of duplicate resources.

### 3.4. Level 3: Manual Service Composition into Cross-Platform Workflows

While the two previous levels refer to search and discovery, the other three levels relate to cross-platform service workflows. The idea is to make use of the respective platforms' specific services to benefit from the best possible workflows as bespoke processing pipelines. The easiest way to realise cross-platform workflows is to develop them manually; this requires knowledge of the APIs and technologies used for each service/tool involved in the workflow and the development of the required wrappers for making them compatible with the workflow execution system.

Figure 9 demonstrates a working example for automated translation from German to Latvian (through English), followed by running the Latvian translation through a dependency parser. If a workflow is developed manually, incompatibilities with regard to data formats are not relevant. Furthermore, regardless of their implementation as server- or client-side code, such workflows could be described as first-class citizens of the respective repository using its metadata scheme (i. e., the workflow gets a name, ID, description etc.) and stored in the repository so that other users can discover, retrieve, potentially modify and apply them.

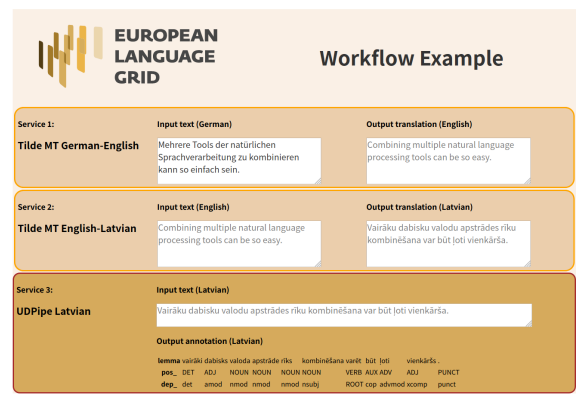

Figure 9: A cross-platform workflow example

A similar approach was implemented in the project Open-MinTeD (OMTD) (Labropoulou et al., 2018) using the Galaxy workflow management system.[10] Three types of LT components are supported: (1) components packaged in Docker images that follow the OMTD specifications; (2) components wrapped with UIMA or GATE, available in a Maven repository; (3) Text and Data Mining web services that run outside the OMTD platform and that follow the OMTD specifications. Each component is registered in the OMTD repository by providing a metadata record. These are curated by the platform administrators and published in the catalogue when the components have been checked for conformity to the OMTD specifications. For each component, a Galaxy wrapper was automatically created from the metadata record and ingested to the Galaxy server. A Galaxy wrapper is an XML file[11] that allows (1) adding the component to the toolbox of the workflow editor and (2) invoking the component. The LT providers or other OMTD users can use Galaxy to chain LT components into workflows, set parameters and publish the workflow. Each processing step is executed as a command line tool within a Docker container in a Mesos cluster.

---

[10]https://galaxyproject.org
[11]https://docs.galaxyproject.org/en/latest/dev/schema.html

| Level | Description | Complexity | What is required from each participating platform? |
|---|---|---|---|
| 1 | Simple cross-platform search | * | Publicly available repository index or repository search API |
| 2 | Complex cross-platform search | ** | Exchange of repository metadata records with shared semantics |
| 3 | Manuel composition of cross-platform workflows | *** | Publicly available service APIs; legal and organisational interoperability |
| 4 | Automated service composition into cross-platform workflows | **** | Publicly available service APIs with complete semantic descriptions |
| 5 | Sophisticated cross-platform development workflows | ***** | Protocols for the automated training and exchange of resources (models etc.) |

Table 2: Five levels of AI/LT platform interoperability (focusing upon service discovery as well as workflow composition)

### 3.5. Level 4: Automated Service Composition into Cross-Platform Workflows

In addition to Level 3, we can foresee a more sophisticated way of composing cross-platform workflows grounded in deep semantic descriptions of the corresponding APIs and data formats. If the workflow manager has access to semantic metadata that describe the services' requirements regarding APIs and data formats, workflows can be partially automated through GUIs that enable their composition. The difference to Level 3 is that the workflow manager, or the different platforms, have access to explicitly represented knowledge that describes which services are interoperable, i. e., the manual mapping of data formats and their attributes or values is not necessary. For this to work, services and workflows need to be first class citizens of the metadata scheme (including persistence, discovery, retrieval, billing etc.); all data formats need to be agreed upon or made interoperable through a shared semantic space.

### 3.6. Level 5: Sophisticated Cross-Platform AI/LT Development Workflows

The last level of platform interoperability relates to fully realised and automated AI/LT development workflows. This scenario enables the automated development of new AI/LT tools by providing fully interoperable data and tool exchange pipelines. For example, an annotated data set available in ELG could be made available to AI4EU by ingesting it into AI4EU's Experiments instance, training a new model and then moving the resulting Kubernetes artefact back into ELG with an automatically pre-filled partial metadata record. As the metadata records are available cross-platform, the resulting new resource is also automatically discoverable through AI4EU's search (Levels 1 and 2).

### 3.7. Legal and Operational Interoperability

In addition to the technical and organisational aspects, which are the main focus of this article, there are the dimensions of legal and operational interoperability, which are equally complex and which also need to be successfully addressed to arrive at full platform interoperability. Here, we can only scratch the surface.

An important aspect relates to authentication and authorisation. Do platforms only expose services and resources that can be freely shared? Can a registered user of platform X, who searches for service A on platform X and finds it in platform Y, use service A in platform Y, in which the user is *not* registered? Technically, this can be solved easily but in order to arrive at a solution that works for all parties and platforms involved, legal interoperability must be reached, i. e., collaboration agreements and policies need to be drawn up and endorsed by all. Legal interoperability also relates to the standard licenses that platforms need to agree upon for sharing different types of digital objects, from data sets to language models to containerised processing services. Especially with regard to commercial services and cross-platform workflows that include such services, policies and mechanisms for billing and brokering need to be agreed upon. For the formal representation of licensing terms and policies, the W3C standard Open Digital Rights Language (ODRL) offers a good solution (Iannella et al., 2018; Iannella and Villata, 2018).

## 4. Conclusions and Next Steps

The interoperability of the AI and LT platforms our community develops is of crucial importance collaboratively to develop something that is, jointly, more useful and more innovative than the sum of its parts. However, achieving platform interoperability requires commitment and effort by all parties involved, i. e., the platform developers need to be cooperative and actually *want* to participate in a wider group of interoperable platforms. To achieve Level 1 interoperability, a participating platform needs to offer a documented and public search API for (parts of) its repository and, for more advanced levels, also access to documented and public APIs for its processing services to enable the manual or automated composition of service workflows (Table 2).

Platform interoperability can be realised on various levels, from simple to highly complex. As an initial roadmap, the authors would like to suggest to the AI/LT community to start implementing platform interoperability at Level 1 and then attempt to realise the various stages up to Level 5. There is a multitude of aspects that can and must be addressed in addition to cross-platform search and cross-platform service workflows, among others, user authentication, shared data storage, shared compute infrastructure as well as shared organisational and legal approaches. An instrument to arrive at joint understanding of shared technical concepts is standardisation, which could include processing APIs and the shared semantic space (vocabulary, location, functionalities etc.). A joint European approach towards platform interoperability could provide a competitive advantage when compared to the very-large-industry-driven developments followed on other continents.

## Acknowledgments

The work presented in this paper has received funding from the European Union's Horizon 2020 research and innovation programme under grant agreements no. 825627 (European Language Grid), no. 825619 (AI4EU), no. 825182 (Prêt-à-LLOD) and no. 780602 (Lynx) as well as from the German Federal Ministry of Education and Research (BMBF) through the project QURATOR (Wachstumskern no. 03WKDA1A) and from the German Federal Ministry for Economic Affairs and Energy (BMWi) through the project SPEAKER (no. 01MK19011). Finally, the authors would like to thank all project teams involved in the different projects for their respective contributions.

## 5. Bibliographical References